\documentclass[10pt,twocolumn,letterpaper]{article}

\usepackage{iccv}
\usepackage{times}
\usepackage{epsfig}
\usepackage{graphicx}
\usepackage{amsmath}
\usepackage{amssymb}
\usepackage[table]{xcolor}
\usepackage{flushend}

\usepackage[pagebackref=true,breaklinks=true,letterpaper=true,colorlinks,bookmarks=false]{hyperref}

\iccvfinalcopy 

\def\iccvPaperID{33} 

\ificcvfinal\fi

\begin{document}

\title{Knowledge-Guided Short-Context Action Anticipation in Human-Centric Videos}

\author{Sarthak Bhagat, Simon Stepputtis, Joseph Campbell, Katia Sycara\\
The Robotics Institute, Carnegie Mellon University\\
{\tt\small \{sarthakb, sstepput, jacambe, sycara\}@andrew.cmu.edu}
}

\maketitle
\ificcvfinal\thispagestyle{empty}\fi

\begin{abstract}
This work focuses on anticipating long-term human actions, particularly using short video segments, which can speed up editing workflows through improved suggestions while fostering creativity by suggesting narratives. To this end, we imbue a transformer network with a symbolic knowledge graph for action anticipation in video segments by boosting certain aspects of the transformer's attention mechanism at run-time. Demonstrated on two benchmark datasets, Breakfast and 50Salads, our approach outperforms current state-of-the-art methods for long-term action anticipation using short video context by up to $9\%$.
\end{abstract}

\section{Introduction}

The ability to predict which actions may happen after a particular video segment ends has multiple use cases in video understanding, including video production and editing \cite{Hutchinson2020VideoAU, Zhu2020ACS, Bhagat2020DisentanglingMF, Wu2021TowardsLV}.
This work focuses on anticipating actions from short video segments and provides potential avenues to enhance the editing process. 
In particular, the ability to extract actions from a video segment can be utilized in two manners: 1) It allows for intelligent clip suggestions for future editing, namely the ability to suggest videos given what will likely happen next, and 2) it provides information on what \textit{generally would happen}, which allows editors to refine their composition to either confirm or contradict a viewer's expectation. As a particular challenge, our work addresses video understanding in the context of short video segments that only span seconds to predict which actions will happen for how long after the end of the segment \cite{Akbarian2017EncouragingLT}. 

\begin{figure}
    \centering
    \includegraphics[width=1.0\linewidth]{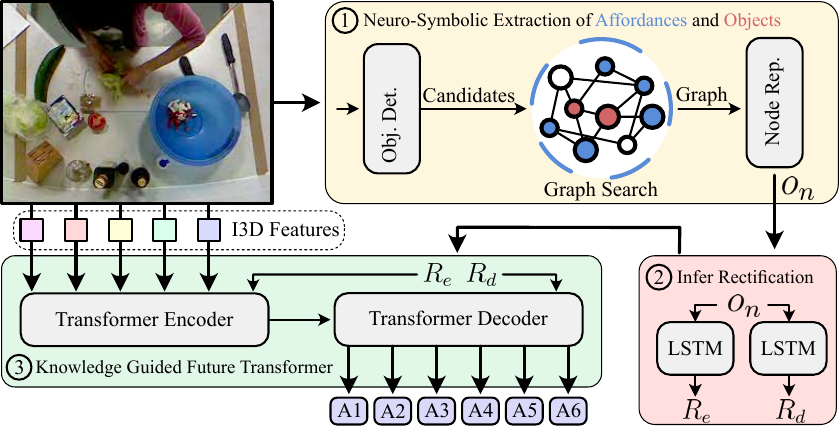}
    \caption{Overview of our proposed Knowledge Guided Action Anticipation approach.}
    \label{fig:overview}
\end{figure}

This work presents a novel approach to action anticipation from video segments by combining symbolic domain knowledge with the video comprehension capabilities of transformer-based architectures \cite{Selva2022VideoTA}. While such transformers often require a long context window to comprehend the underlying scene \cite{9587279}, extracting the information about relevant objects and associating them with the possible set of actions that could be taken with them enables us to make inferences about future actions even with little context. 
We propose to augment the attention mechanism \cite{mha} of transformer-based action-anticipation architectures by utilizing symbolic domain knowledge to boost or suppress the attention given to various features presented in the video. 
This allows us to predict future actions accurately, particularly from short-horizon observations -- a key aspect that prior works \cite{farha2018, farha2019, ke2019, sener2020, farha2021, gong2022future} in action anticipation fail to cater to. 

In our work, we utilize Knowledge Graphs (KG) to capture the relationship between entities present in the video and link them to their respective affordances and the potential tools that could be used to afford them in a particular way. 
Prior work \cite{yuke_2014, 7102751, 10.5555/3524938.3525424, Ghosh2022TextDerivedKH} has introduced efficient methods of identifying such relationships, which can subsequently be utilized to identify the potential for certain actions. For example, in a kitchen scene, seeing a tomato and a knife can enable the potential action of cutting the tomato as a tomato has the affordance of \textit{can be cut} while a knife affords the ability to \textit{cut}. Intuitively, we utilize such knowledge to alter the attention mechanism of transformer-based action-anticipation methods to re-focus what features the transformer pays attention to when predicting the list and duration of future actions. 

We demonstrate our method on two common long-term action-anticipation benchmarks, namely the 50Salads \cite{50salads} and Breakfast \cite{Kuehne_2014_CVPR} datasets, and show superior performance as compared to current state-of-the-art methods.

\section{Related Works}

\begin{figure*}
    \centering
    \includegraphics[width=1\textwidth]{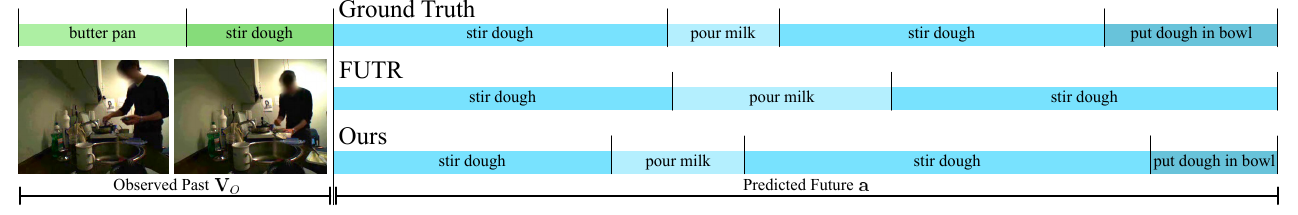}
    \caption{Example of our method on the \textit{Breakfast} dataset, observing $10\%$ (green hues) and predicting another $30\%$ (blue hues).}
    \label{fig:segmentation}
\end{figure*}


\textbf{Concept Learning.}
The emerging field of relevant concept extraction from visual inputs involves identifying and extracting relevant visual and non-visual entities from the input data \cite{NIPS2013_d2ed45a5, CBM2020, Kim2020VisualCR, cao2021concept, mei2022falcon}.
\cite{marino} introduced a knowledge graph as a structured prior for image classification and proposed the Graph Search Neural Network, demonstrating its performance improvement by integrating knowledge graphs into the vision classification pipeline. Further, \cite{Bhagat2023SampleEfficientLO} extended it to include the augmentation of novel concepts, encompassing visual objects and compound concepts such as affordances, attributes, and scenes.
In this work, we extend the idea by refining the propagation framework from \cite{Bhagat2023SampleEfficientLO} to identify relevant object affordances. We then utilize these concepts to predict a possible set of actions in a video based on a short previous context.

\textbf{Action Anticipation.}
The task of action anticipation from videos \cite{HU2022395} revolves around predicting future actions based on a specific segment of the video. With recent advancements in foundational vision models and the availability of large-scale human-centric datasets \cite{Damen2018EPICKITCHENS, Damen2022RESCALING}, this domain has gained significant attention. Many recent approaches have been developed to predict a single future action within a short time frame, typically spanning a few seconds \cite{furnari2019, gammulle2019, miech2019, sener2019, sener2020, fernando2021, girdhar2021, roy2021}. However, a notable emerging trend is long-term action anticipation, which emphasizes predicting a sequence of future actions occurring in the distant future from a lengthy video \cite{farha2018, farha2019, ke2019, sener2020, farha2021, gong2022future}.
While much attention has been paid to predicting long-term actions with ample video context, limited research has addressed using short video contexts to predict long-term future action sequences. Particularly in this challenging domain, a deep understanding of the scene objects and their contextual utilization can be helpful. 

\section{Knowledge Guided Action Anticipation}

In this section, we introduce the two main components of our approach: 1) a transformer-based architecture to extract future actions based on~\cite{gong2022future} (see Section~\ref{sec:futr}) and 2) a neuro-symbolic architecture to extract affordances from video frames (see Section~\ref{sec:kgextraction}). Subsequently, in Section~\ref{sec:correction}, we introduce how the domain knowledge from our neuro-symbolic pipeline can be utilized to re-focus the attention of the video transformer by introducing a correction matrix to its attention mechanism.

\subsection{Problem Setup}
Given an observed video segment $\mathbf{V}_O$, we create a policy $\mathbf{a} = \pi(\mathbf{V}_O)$ that predicts a sequence of actions $\mathbf{a}$ happening after the end of $\mathbf{V}_O$. The video sequence $\mathbf{V}_O \in \mathcal{R}^{H \times W \times C \times N}$ is represented as a four-dimensional matrix describing the height $H$, width $W$, and channels $C$ of each video frame $\mathbf{V}_i$, and the number of frames $N$.

We train our model from a dataset $\mathcal{D} = [\mathbf{s}_n, \dots, \mathbf{s}_N]$ where each sample $\mathbf{s}_n = [\mathbf{V}_n, a_n]$ contains the video-frame $\mathbf{V_n}$ and action-label $\mathbf{a}_n$, where $n$ is the frame index. After training, we provide the trained policy $\pi$ with a new, previously unseen video sequence showing $\alpha$ percent of the full video, tasked with predicting the most likely action for each frame in the following $\beta$ percent of the remaining video. 

Figure~\ref{fig:overview} demonstrates the main components of our approach across our three-step process. In the following sections, we will detail the Future Transformer (FUTR)~\cite{gong2022future} (Section~\ref{sec:futr}), and how we extract symbolic knowledge for a given video segment using a neuro-symbolic architecture~\cite{Bhagat2023SampleEfficientLO} (Section~\ref{sec:kgextraction}). Subsequently, we will detail how transformer architecture can be enhanced by imbuing it with symbolic knowledge (Section~\ref{sec:correction}).


\subsection{Future Transformer (FUTR)}
\label{sec:futr}


Future Transformer (FUTR) \cite{gong2022future} is a long-term action anticipation transformer architecture that consists of a transformer encoder and a transformer decoder.
The encoder is responsible for processing visual features extracted from the observed segment of a video by employing multi-head self-attention \cite{mha}.
To accomplish this, the encoder, $f_e$, employs a stack of $L_e$ layers, each containing multi-head attention mechanisms, layer normalization \cite{Ba2016LayerN}, and feed-forward networks, all interconnected through residual connections \cite{He2015DeepRL}, i.e., $\mathbf{x}_n^{L_e} = f_e(\mathbf{x}_n^0)$ for $\mathbf{x}^0_n = \mathbf{W}_f \mathbf{x}_n^f \in \mathcal{R}^D$, where $\mathbf{x}_n^f$ are the I3D \cite{8099985} features for the $n^{th}$ frame of the observed video $\mathbf{V}_O$, $D$ represents the encoding dimension, and $\mathbf{W}_f$ is a learnable weight matrix. 
The resulting output is then provided to a classifier, $f_{obs}$, which is a fully connected layer followed by a softmax activation function, determining the actions corresponding to the observed part of the video segment: $\mathbf{a}_{obs} = f_{obs}(\mathbf{x}_n^{L_e})$.


The decoder employs a cross-attention mechanism that uses the embeddings of the observed sequence generated by the encoder. After processing through $L_d$ layers of the decoder, architecturally similar to the encoding ones, the output, $\mathbf{q}_n^{L_d}$, is obtained, i.e., $\mathbf{q}_n^{L_d} = f_d(\mathbf{x}_n^{L_e})$.
Subsequently, we utilize two separate, fully connected networks for predicting the future actions $\mathbf{a}_{pred}$ and their durations $\mathbf{d}_{pred}$ respectively.
\begin{equation}
    \mathbf{d}_{pred} = f_{dur}(\mathbf{q}_n^{L_d}) \hspace{0.3cm} \text{and} \hspace{0.3cm} \mathbf{a}_{pred} = f_{act}(\mathbf{q}_n^{L_d})
\end{equation}

Putting both these components together, long-term relationships between past and future actions are learned via this transformer framework. The loss for the transformer is comprised of the framewise cross-entropy loss for the observed action sequence combined with the L2 and cross-entropy loss over the durations and actions of the predicted action sequence, respectively.
With a defined methodology for predicting future actions from video segments in place, the subsequent sections delve into the process of enhancing it with symbolic domain knowledge. 


\setlength{\tabcolsep}{3pt}
\begin{table*}
\centering
\footnotesize
\begin{tabular}{c|cccccccc|ccc}
\hline
& \multicolumn{8}{c}{Frame-wise ($\uparrow$ larger is better) / Action Sequence ($\downarrow$ smaller is better)} & \multicolumn{2}{|c}{Next Action ($\uparrow$)}\\\
Model & 5-10 & 5-20 & 5-30 & 5-50 & 10-10 & 10-20 & 10-30 & 10-50 & 5 & 10 \\
\hline
\multicolumn{11}{c}{50Salads} \\
\hline
KG Baseline \cite{Bhagat2023SampleEfficientLO} & 6.92 / 4.88 & 6.21 / 6.20 & 6.01 / 7.44 & 5.58 / 7.92 & 7.13 / 4.50 & 6.48 / 5.98 & 6.07 / 7.37 & 5.78 / 7.88 & 8.0 & 9.0 \\
Video-Llama \cite{damonlpsg2023videollama} & - / 6.44 & - / 7.20 & - / 7.90 & - / 9.12 & - / 6.12 & - / 6.80 & - / 7.86 & - / 9.02 & 6.0 & 7.0 \\
FUTR \cite{gong2022future} & 8.90 / 2.98 & 7.46 / 4.52 & 7.29 / 5.40 & 8.63 / 6.80 & 15.17 / 2.74 & 11.34 / 4.04 & 11.31 / 4.98 & 11.36 / 6.04 & 12.0 & 36.0 \\
Ours & \textbf{17.86} / \textbf{2.84} & \textbf{16.25} / \textbf{4.22} & \textbf{10.84} / \textbf{5.14} & \textbf{9.38} / \textbf{6.70} & \textbf{23.15} / \textbf{2.54} & \textbf{17.28} / \textbf{3.78} & \textbf{16.62} / \textbf{4.76} & \textbf{13.61} / \textbf{5.74} & \textbf{14.0} & \textbf{42.0}\\
\hline
\multicolumn{11}{c}{Breakfast} \\
\hline
KG Baseline \cite{Bhagat2023SampleEfficientLO} & 5.44 / 8.22 & 4.95 / 9.10 & 4.22 / 9.66 & 3.98 / 10.02 & 6.02 / 7.90 & 5.15 / 8.77 & 4.86 / 9.21 & 4.51 / 9.78 & 7.22 & 12.31 \\
Video-Llama \cite{damonlpsg2023videollama} & - / 11.20 & - / 12.24 & - / 13.62 & - / 13.82 & - / 11.08 & - / 12.04 & - / 12.98 & - / 13.22 & 5.39 & 9.80 \\
FUTR \cite{gong2022future} & 9.54 / 1.63 & 7.24 / 2.07 & 6.42 / 2.40 & 5.58 / 3.02 & 14.70 / \textbf{1.41} & 12.55 / \textbf{1.76} & 12.10 / \textbf{2.06} & 11.71 / \textbf{2.62} & 23.97 & \textbf{30.05} \\
Ours & \textbf{9.91} / \textbf{1.60} & \textbf{7.95} / \textbf{2.02} & \textbf{6.86} / \textbf{2.34} & \textbf{5.88} / \textbf{2.98} & \textbf{15.53} / \textbf{1.41} & \textbf{13.52} / \textbf{1.76} & \textbf{13.07} / 2.09 & \textbf{11.94} / 2.63 & \textbf{25.25} & 26.45 \\
\hline
\end{tabular}
\caption{Performance of the proposed approach compared to the current SOTA for different values of $\alpha$ and $\beta$ percent (top row). 
We also compare to Video-Llama, a multimodal fusion model utilizing large language models, on the action-sequence predictions task. 
}
\label{quant}
\end{table*}

\subsection{Visual Concept Extraction}
\label{sec:kgextraction}

To extract objects and affordances from video frames, we utilize~\cite{Bhagat2023SampleEfficientLO}, extracting these features by utilizing a domain-specific knowledge graph. 
Intuitively, this approach utilizes a neural object detector to extract a set of initial objects and subsequently utilizes them to initialize the knowledge graph $\mathcal{K}$. 
We created the graph $\mathcal{K}$ consisting of two types of nodes: object nodes (e.g., \textit{salt}, \textit{knife}, \textit{bowl}) and affordance nodes (e.g., \textit{graspable}, \textit{pourable}, \textit{cuttable}). The knowledge graph connects each object to its respective affordance, and each affordance is linked to its corresponding tool. For example, a \textit{tomato} has a connection to \textit{cuttable}, which, in turn, connects to \textit{knife}.
Then, we employ a graph-search approach to propagate information from the initial nodes to relevant connected nodes throughout an iterative process. Finally, we generate a latent representation of all relevant nodes $\mathbf{o}_n$. 

Consider, $\mathbf{c}^k_A$ and $\mathbf{c}^k_C$ represent the active and the candidate nodes respectively at the $k^{th}$ propagation step, for $k \in \{0, 1, ..., T\}$.
We start by running an object detection algorithm \cite{liu2023grounding}, $f_{OD}$, to identify objects in the scene for the $n^{th}$ frame. These objects serve as the initial nodes in our graph for the propagation process, so, $\mathbf{c}^0_A = f_{OD}(\mathbf{V_n})$. At each step of propagation, the neighbors of the previously active nodes in $\mathcal{K}$ are considered candidate nodes.
The importance network, $f_I$, computes the importance of each candidate node, conditioned on the visual input, and finally, the ones above a certain importance threshold, $\gamma$, are expanded.

\begin{equation}
    \mathbf{c}^k_A = \{ \mathbf{c}^k_C \mid f_I(\mathbf{c}^k_C) > \gamma \}
\end{equation}

Through a series of $T$ propagations, we determine the associated affordances for each object in the scene and the potential tools that can be used for those affordances. For example, if a \textit{tomato} is detected, we associate it with the affordance of being \textit{cuttable} and link it to the tool \textit{knife} for cutting. This helps us understand the range of actions possible in this setting. Each identified object, affordance, and the respective tool to perform this affordance is then passed onto the context network, $f_C$, to obtain the vectors corresponding to each finally active node in the KG, i.e., $\mathbf{o}_n = f_C(\mathbf{c}^T_A)$. The collection of these context vectors, $\mathbf{o}_n$, is then utilized to modify the attention weights in the following steps of our approach. 

\subsection{Knowledge Guided Attention Mechanism}
\label{sec:correction}

In this section, we introduce our main contribution, which is a methodology for augmenting the attention mechanism of transformers with symbolic knowledge contained within a KG.
%
%
Specifically, we augment FUTR with a KG that contains various domain-specific commonsense relations that enable our model to identify the objects present in the scene, establish connections with their respective affordances, and consequently identify the most plausible set of actions that can be performed in the current context. This information is used to modify the weight of each visual feature in the attention.

Having identified the set of concepts in the scene, we can leverage them to adjust our attention weights. This modification allows our model to prioritize the features associated with objects having relevant affordances, giving them higher importance compared to those not present in the scene. We obtain a separate knowledge-guided rectification matrix for our encoder and decoder, namely $\mathbf{R}_e$ and $\mathbf{R}_d$ respectively. This is obtained by processing the context vectors, $\mathbf{o}_n$, through our LSTM-based knowledge-guided rectification functions, $\mathbf{R}_{e/d} = f^{e/d}_{KG}(\mathbf{o}_n)$. This function takes in the zero-padded context vectors and outputs a matrix that signifies the weight we need to assign to each visual feature. 
Therefore, we modify the standard multi-head attention \cite{mha} to obtain our knowledge-guided attention separately for our transformer encoder and decoder.
\begin{equation}
    \text{KG-Attn}_{e/d}(\mathbf{Q}, \mathbf{K}, \mathbf{V}) = softmax\big(\frac{\mathbf{Q} \mathbf{R}_{e/d} \mathbf{K}^T}{\sqrt{d_k}}\big)\mathbf{V}
\end{equation}

These rectification matrices modify the weights of attention so that features corresponding to the objects with relevant affordances are weighted according to the likelihood of an action being performed with them.

\section{Experiments}
In this section, we present the assessment of our proposed neuro-symbolic action anticipation method in comparison to the existing benchmark utilized for this task.

\textbf{Datasets.}
We evaluate the effectiveness of our approach using two publicly available benchmark datasets for action anticipation in kitchen-based videos: the 50Salads dataset \cite{50salads} with its five splits, densely annotated with 17 fine-grained action labels and 3 high-level activities, and the Breakfast dataset \cite{Kuehne_2014_CVPR} with four splits, categorizing each frame into one of 10 breakfast-related activities using a comprehensive set of 48 fine-grained action labels.

\textbf{Metrics.}
To evaluate the efficacy of our approach, we calculate the mean over classes (MoC) accuracy. This metric is computed by comparing the predicted actions to the ground-truth actions for all future frames within the horizon window.
To quantify the ability of our model to identify the sequence of next actions without considering their durations, we employ a metric that computes the minimum number of addition, deletion, or substitution operations required to exactly match the predicted and the ground truth order action sequence.
Finally, we also employ immediate single next-action prediction as a metric.

\textbf{Quantitative Evaluation.}
We compare our approach against a KG-only baseline \cite{Bhagat2023SampleEfficientLO}, an LLM-based baseline \cite{damonlpsg2023videollama}, and the current state-of-the-art in long-term action anticipation \cite{gong2022future}, where each metric is averaged over all the splits. The details about the implementation of these baselines are available in the supplementary.
As can be viewed in Table \ref{quant}, our approach outperforms the current state-of-the-art in long-term action anticipation using short context in all the metrics on the 50Salads dataset and on seven out of the ten metrics we used on the Breakfast dataset. On the MoC metric, we outperform the baseline by up to $9\%$ on 50Salads and $1\%$ on Breakfast. 


\textbf{Qualitative Evaluation.}
\begin{figure}
    \centering
    \includegraphics[width=0.3\linewidth]{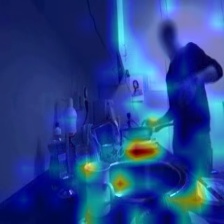}
    \includegraphics[width=0.3\linewidth]{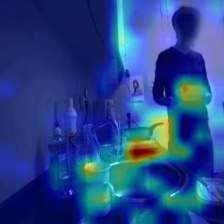}
    \includegraphics[width=0.3\linewidth]{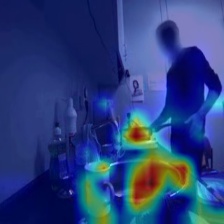}
    \caption{Heatmaps highlighting visual features where the attention is boosted as a result of our knowledge-guided rectification.}
    \label{fig:heatmaps}
\end{figure}
We also showcase an example to compare our approach against \cite{gong2022future} by looking at the time-series segmentation of the predicted future actions.
Figure \ref{fig:segmentation} depicts an example where the model observes two actions in the $10\%$ observed segment of the video and then tries to predict what actions take place in the next $30\%$ of the video. While our model identifies all four actions along with their durations accurately, the baseline approach fails to identify the last action of \textit{pour the dough into the bowl}. This can be attributed to the ability of our approach to focus on the set of objects present in the scene (specifically, the \textit{bowl}), associate them with their respective affordances (of being able to \textit{contain} the dough) and therefore, identify the possible set of actions that could be taken in the future (i.e, \textit{pouring the dough in the bowl}). 

As illustrated in heatmaps depicted in Figure \ref{fig:heatmaps} (right), our approach demonstrates the capability to focus not only on currently utilized objects but also on those with potential for future use. 
During the execution of the \textit{stirring the dough} action, the approach enhances features corresponding to both the \textit{pan} and the \textit{bowl}. This specific instance is intriguing as our approach accurately recognizes that the subsequent action after \textit{stirring the dough} would likely involve \textit{placing the dough into the bowl}.



\section{Conclusion}

Our novel knowledge-guided action anticipation approach considers both objects and their affordances in the scene, augmenting it with commonsense relations via a KG. By supplementing the action anticipation transformer with a KG, our methodology outperforms the current state-of-the-art long-term action anticipation in videos with short video contexts, establishing a new benchmark for this task.

\paragraph{Acknowledgements.}
We would like to acknowledge the support from DARPA under grant HR001120C0036, AFOSR under grants FA9550-18-1-0251 and FA9550-18-1-0097, and ARL under grant W911NF-19-2-0146 and W911NF-2320007.



{\small
\bibliographystyle{ieee_fullname}
\bibliography{main}
}

\end{document}